# Node2Vec-DGI-EL: A Hierarchical Graph Representation Learning Model for Ingredient-Disease Association Prediction

Leifeng Zhang[1†], Xin Dong[123†], Shuaibing Jia[1], Jianhua Zhang[13*]

1. Medical Engineering Technology and Data Mining Institute, Zhengzhou University, 450000, 100 Science Avenue, Zhengzhou, China
2. Faculty of Innovation Engineering, Macau University of Science and Technology, 999078, Avenida Wai Long, Taipa, Macao, China
3. School of Pharmacy, Macau University of Science and Technology, 999078, Avenida Wai Long, Taipa, Macao, China

† These authors contributed equally to this work.
*Corresponding author: Jianhua Zhang. E-mails: petermails@zzu.edu.cn,

# Abstract

Traditional Chinese medicine, as an essential component of traditional medicine, contains active ingredients that serve as a crucial source for modern drug development, holding immense therapeutic potential and development value. A multi-layered and complex network is formed from Chinese medicine to diseases and used to predict the potential associations between Chinese medicine ingredients and diseases.

This study proposes an ingredient-disease association prediction model (Node2Vec-DGI-EL) based on hierarchical graph representation learning. First, the model uses the Node2Vec algorithm to extract node embedding vectors from the network as the initial features of the nodes. Next, the network nodes are deeply represented and learned using the DGI algorithm to enhance the model's expressive power. To improve prediction accuracy and robustness, an ensemble learning method is incorporated to achieve more accurate ingredient-disease association predictions. The effectiveness of the model is then evaluated through a series of theoretical verifications.

The results demonstrated that the proposed model significantly outperformed existing methods, achieving an AUC of 0.9987 and an AUPR of 0.9545, thereby indicating superior predictive capability. Ablation experiments further revealed the contribution and importance of each module. Additionally, case studies explored potential associations, such as triptonide with hypertensive retinopathy and methyl ursolate with colorectal cancer. Molecular docking experiments validated these findings, showing the triptonide-PGR interaction and the methyl ursolate-NFE2L2 interaction can bind stable.

In conclusion, the Node2Vec-DGI-EL model focuses on TCM datasets and effectively predicts ingredient-disease associations, overcoming the reliance on node semantic information.

## Intorduction

As a vital ingredient of China's traditional medical system, Traditional Chinese Medicine (TCM) boasts a long history and a unique theoretical framework, emphasizing synergistic therapeutic effects through multi-ingredient, multi-target, and multi-pathway mechanisms. However, the complexity of TCM systems has long hindered systematic analysis. The rise of network pharmacology has provided a new perspective for TCM research by constructing multi-layered herb-disease networks to systematically reveal the material basis of efficacy and molecular mechanisms [1]. In recent years, the rapid development of artificial intelligence (AI) and knowledge graph technologies has further accelerated progress in TCM research based on biological networks and big data [2].

Neural network algorithms have emerged as powerful tools for predicting drug-disease associations, significantly advancing drug discovery and repositioning efforts [3]. However, their application in predicting TCM ingredient-disease associations remains underexplored. In conventional drug-disease prediction, Gottlieb et al. [4] employed a logistic regression framework to rank associations using drug-drug and disease-disease similarity metrics. Similarly, Kim et al. [5] predicted pharmacological effects of herbal compounds by integrating multiple similarity measures. Luo et al. [6] adopted a heterogeneous network and random walk model for candidate disease prediction, while their DRRS system [7] utilized the SVT algorithm for drug repositioning. Zhang et al. [8] proposed the SCMFDD algorithm, which projects associations into a low-rank space constrained by drug and disease similarities. More recently, Yu et al. [9] introduced the LAGCN algorithm leveraging graph convolution and attention mechanisms, and Kang et al. [10] combined BERT-based semantic extraction with graph convolution for association prediction.

Despite these advances, existing methods predominantly rely on constructing drug-drug and disease-disease similarity matrices. Consequently, they face critical limitations in TCM ingredient-disease prediction due to the vast scale of herb-ingredient-target-disease networks and the lack of prior knowledge for many nodes, making similarity matrix construction impractical. Although Wang et al. [11] addressed this issue in part with HTINet—using random walks and shallow neural networks to learn node embeddings without semantic dependencies—their method focused solely on herb-target interactions. Subsequently, Duan et al. [12] improved HTINet2 by incorporating residual GCNs and BPR loss optimization, yet it still ignored interactions with other network entities. Therefore, a more robust framework is urgently needed to fully exploit TCM network data for ingredient-disease association prediction.

To address the above issues, we propose the Node2Vec-DGI-EL model for the "Herb-Ingredient-Target-Disease" (HITD) association network, which combines the Node2Vec algorithm [13], the DGI algorithm [14], and an ensemble learning method, with three key advantages. First, the node embeddings generated by Node2Vec provide high-quality initial features for DGI, effectively alleviating the performance degradation caused by the lack of prior knowledge. Second, DGI's global comparison mechanism complements Node2Vec's local structural capture capability, enabling more comprehensive network representation learning and generating high-quality embeddings. Finally, by inputting the embeddings into an ensemble learning model for binary classification, the model mitigates the negative impact of class imbalance on prediction accuracy, thereby improving overall performance. In summary, our approach not only advances the modernization of TCM but also provides theoretical and practical value for drug repurposing and development.

## 2. Materials and methods

### 2.1 Data Collection

This study utilized the Encyclopedia of Traditional Chinese Medicine (ETCM) [15] to collect association data for four key entities: TCMs, active ingredients, protein targets, and diseases. To enhance the biological relevance of the network, we further integrated protein-protein interaction (PPI) data from the STRING [16], applying stringent filtering criteria to retain only high-confidence interactions with a combined score ≥ 700.

The integrated dataset was used to construct a comprehensive "Herb-Ingredient-Target-Disease" (HITD) association network, which encompasses four types of entities (herbs, ingredients, targets, and diseases) and seven types of associations:
- Herb-ingredient associations
- Ingredient-target associations
- Target-disease associations
- Target-target associations
- Herb-disease associations
- Ingredient-disease associations
- Disease-disease associations

The complete network architecture is presented in Table 1.

Table 1 HITD network data statistics.

| Type | Number of nodes | Number of nodes | Number of edges | Source of data |
|---|---|---|---|---|
| Herb-Ingredient | 402 | 6971 | 10304 | ETCM |
| Herb-Target | 399 | 1752 | 50061 | ETCM |
| Herb-Disease | 395 | 2704 | 248845 | ETCM |
| Ingredient-Target | 4185 | 1635 | 75112 | ETCM |
| Ingredient-Disease | 4029 | 2661 | 617530 | ETCM |
| Disease-Target | 4289 | 4887 | 73540 | ETCM |
| Target-Target | 16201 | 16201 | 236930 | STRING |

### 2.2 Model architecture design

The architecture of the model in this study consists of three modules: the node feature extraction module, the node pair feature construction module, and the ensemble learning module, as shown in Figure 1.

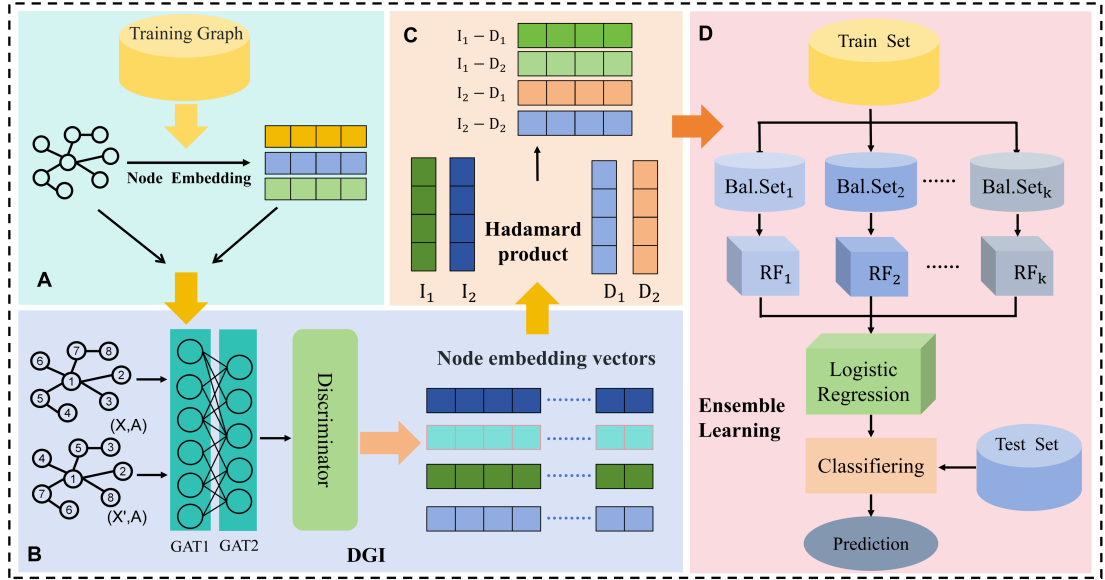

Figure 1 Overall architecture of Node2Vec-DGI-EL. A and B are the node feature extraction modules. C is the node pair feature construction module. D is the ensemble learning module.

### 2.2.1 Node feature extraction modules

This module combined the Node2vec algorithm and the DGI algorithm to extract node features from the graph network. The Node2vec algorithm integrates the ideas of depth-first search (DFS) and breadth-first search (BFS) and introduces two parameters: the return parameter p and the in-out parameter q to control the random walk strategy. It then uses the Skip-Gram model to learn node embedding vectors.

The loss function of the Node2vec algorithm is as follows:

$$L_{\text{Node2Vec}} = \sum_{t=1}^{T} \sum_{-k \leq j \leq k, j \neq 0} \log \sigma(V_{v_t}^T V_{v_{t+j}}) - \sum_{-k \leq j \leq k, j \neq 0} E_{v' \sim P_n(v)} \log \sigma(-V_{v_t}^T V')$$

Where $\sigma$ is the Sigmoid activation function. $V_{v_t}$ is embedding vector for the node $v_t$. $V_{V_{t+j}}$ is embedding vector for the context node $v_{t+j}$. $V'$ is embedding vector of negative node $v'$.

Based on the initial node features captured from the network by Node2vec, the DGI algorithm was then used to further learn deeper node embedding representations. DGI is an unsupervised graph learning algorithm that learns node embeddings by maximizing the mutual information between node representations and global graph representations.

The DGI framework mainly includes four aspects. First, a graph neural network (GAT) [17] was employed to encode each node, generating node embedding representations. Second, the node embeddings were aggregated (mean pooling was used in this paper) to generate a global graph representation. Then, negative samples were generated by disrupting the original graph's node feature order. Finally, through the contrastive learning mechanism, the mutual information between node embeddings and global graph representations was maximized, obtaining high-quality node embeddings.

Global graph representation:
$$\vec{s} = R(h_i{}_{i=1}^{N})$$

Discriminator:
$$D(\vec{h}_i, \vec{s}) = \sigma(\vec{h}_i^T W \vec{s})$$

The loss function of the DGI algorithm is as follows:

$$L_{DGI} = \frac{1}{N+M} \left( \sum_{i=1}^{N} E_{(X,A)} \left[ log\, D\left(\vec{h}_i, \vec{s}\right) \right] + \sum_{j=1}^{M} E_{(\tilde{X},\tilde{A})} \left[ log\left(1 - D\left(\vec{h}_j, \vec{s}\right)\right) \right] \right)$$

Where $\vec{h}_i$ is the embedding representation of the i node, $\vec{s}$ is the global graph representation, N is represents the number of positive sample pairs, while M is represents the number of negative sample pairs.

### 2.2.2 Node pair feature construction module

In order to capture the interactive information between node pairs, common construction methods include concatenation, addition, subtraction, and the Hadamard product. Among these, the Hadamard product is one of the most commonly used methods for constructing node pair feature interactions. It captures similarities and differences between nodes through element-wise multiplication and is both computationally efficient and highly versatile [18][19]. This study selected the Hadamard product to construct node pair features. For nodes i and j, their embedding vectors are denoted as $u_i$ and $u_j$, respectively, and the calculation method for the node pair feature vector is as follows:

$$u_{i,j} = u_i \odot u_j$$

### 2.2.3 Ensemble learning module

Due to the sparsity of the network, the number of negative samples in the node pair feature vector dataset is larger than the number of positive samples. This class imbalance issue can lead to machine learning models being biased towards the majority class, affecting the accuracy of predictions [20]. Ensemble learning has shown better performance in improving the classification performance on imbalanced datasets [21]. Therefore, this study adopted the stacking ensemble learning model [22], as shown in Figure 1-C. First, the negative samples were divided into multiple independent subsets, each having the same number of samples as the positive class, and each negative subset was combined with the positive samples to construct multiple balanced subsets (Balanced Subset, abbreviated as: Bal.Set). Then, for each balanced subset, the Random Forest algorithm (RF) [23] was used as the base learner for training. Finally, the Logistic Regression algorithm was used as the meta-learner to integrate the predictions from the base learners, generating a more robust prediction result.

### 2.3 Model training and evaluation

### 2.3.1 Data division

To study the potential associations of ingredients and diseases, this study randomly hid 30% of known ingredient-disease node pairs in the complete HITD network, leaving the rest of the structure unchanged for training. By obtaining the node embedding vector, the hidden edge node pair was calculated as a positive sample (label 1), and negative samples were sampled proportionally from the unconnected ingredient-disease node pairs (label 0) to construct the feature vector dataset. The dataset was divided into training and test sets at a 7:3 ratio for training and evaluation of the ensemble learning model.

### 2.3.2 Evaluation Methods

The ingredient-disease association prediction task can be regarded as a typical binary classification problem[24]. This paper uses two threshold-independent evaluation indicators [25,26], as well as four threshold-dependent evaluation indicators: Accuracy, Precision, Recall, and F1 score to comprehensively evaluate the model's performance. In the model superiority evaluation, parameter tuning was first performed to analyze the impact of different hyperparameters on model performance, and the most suitable parameter combination was selected. In addition, the model was compared with several existing models that have received widespread attention in recent years. The ablation experiment explored the importance of each module in the model. Finally, ingredient-disease association edges were deleted at different proportions (ranging from 10% to 90%), and robustness experiments were conducted to validate the model's stability under data missing conditions.

### 2.3.3 Case study verification method

This study obtained ingredient targets based on SwissTargetPrediction[27] and the SEA[28], and screened disease targets using GeneCards[29] (relevance score > median). Intersection targets were obtained using the Venny tool, and a PPI network was constructed with the STRING. The network was visualized using Cytoscape 3.9.1[30], and, combined with the cytoHubba[31] plugin, the top 10 core targets were screened using the MCC algorithm. Ingredient and target protein structures were obtained from PubChem[32] and the PDB[33]. After conversion to PDBQT format using Open Babel[34], molecular docking was performed with AutoDock 1.5.7[35], and the docking results were finally visualized using PyMOL[36].

## 3 Results

### 3.1 Model selection

This section compares five classic graph representation learning algorithms, including DeepWalk[37], Node2Vec, Struc2vec[38], LINE[39], and SDNE[40], combined with the DGI algorithm (Table 2). The results show that Node2Vec-DGI-EL demonstrates the optimal predictive performance, with all indicators outperforming other models, achieving an AUC value of 0.9976 and an AUPR value of 0.9398. Furthermore, for this model, the impact of different embedding dimensions of the Node2Vec algorithm on model performance is explored (Table 3). The results indicate that when the Node2Vec algorithm uses an embedding dimension of 128, the model's prediction results are optimal.

Table 2 The prediction results of five combination models.

| DeepWalk-DGI-EL | 0.9969 | 0.9329 | 0.9693 | 0.6993 | 0.9744 | 0.7756 |
| Struc2vec-DGI-EL | 0.9972 | 0.9257 | 0.9717 | 0.7089 | 0.9753 | 0.7857 |
| LINE-DGI-EL | 0.9971 | 0.9171 | 0.9694 | 0.6997 | 0.9753 | 0.7762 |
| SDNE-DGI-EL | 0.9971 | 0.9182 | 0.9718 | 0.7096 | 0.9759 | 0.7865 |
| Node2Vec-DGI-EL | **0.9976** | **0.9398** | **0.9752** | **0.7254** | **0.9777** | **0.8026** |

Table 3 The prediction results of different embedding dimensions of Node2Vec.

| Dimensions | AUC | AUPR | ACC | Precision | Recall | F1 |
|---|---|---|---|---|---|---|
| 64 | 0.9954 | 0.9379 | 0.9699 | 0.7124 | 0.9732 | 0.7889 |
| **128** | **0.9976** | **0.9398** | **0.9752** | **0.7254** | **0.9777** | **0.8026** |
| 192 | 0.9959 | 0.9382 | 0.9691 | 0.7175 | 0.9723 | 0.7842 |
| 256 | 0.9965 | 0.9391 | 0.9706 | 0.7169 | 0.9726 | 0.7928 |
| 320 | 0.9954 | 0.937 | 0.9709 | 0.7185 | 0.9723 | 0.7941 |

## 3.2 Parameter tuning

Based on the 128-dimensional embedding of the Node2Vec algorithm, this section analyzed the key hyperparameters of the DGI algorithm in the model, including learning rate, output dimension, hidden layer dimension, and attention head count. By comparing the model performance under different parameter settings and using AUPR as the evaluation metric, the impact of each parameter on the model's performance was assessed. The experimental configurations are detailed in Table 4.

The results show that the optimal parameters of the DGI algorithm are as follows: the learning rate is 0.001; the hidden layer dimension is 192; the output layer dimension is 128; the attention head count is 4.

Table 4 Hyperparameter Search Space and Analysis Figures.

| Hyperparameters | Tested Values | Analysis Figure |
|---|---|---|
| Learning Rate | 0.0001, 0.0005, 0.001, 0.005, 0.01, 0.05 | Figure 2-A |
| Hidden Layer Dimension | 128, 192, 256, 320, 384, 448 | Figure 2-B |
| Output Dimension | 64, 96, 128, 160, 192, 224 | Figure 2-C |
| Attention Head Count | 2, 4, 6, 8, 10, 12 | Figure 2-D |

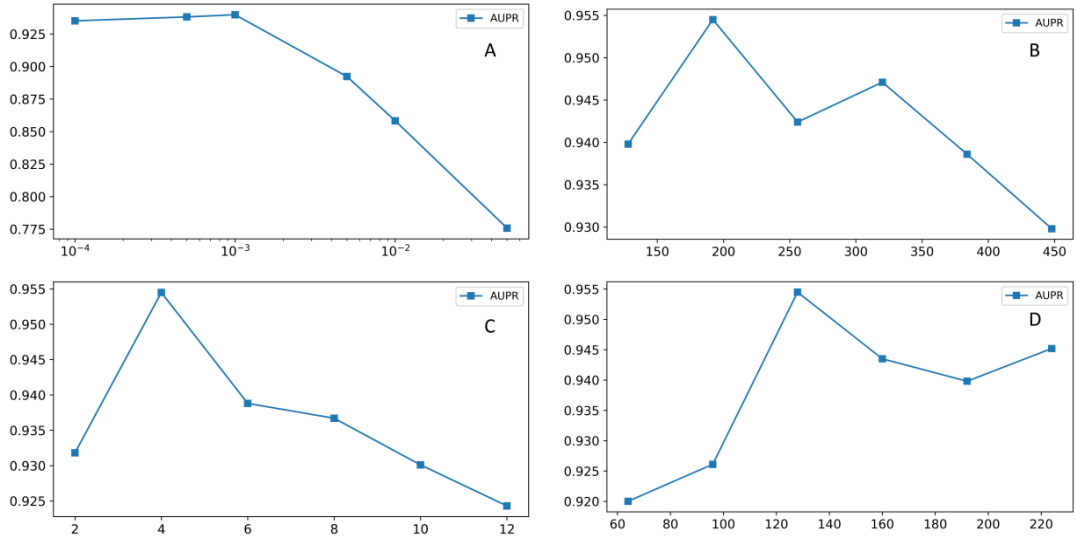

Figure 2. Impact of hyperparameter configurations on model performance. A: Learning rate.B: Hidden layer dimension.C: Number of attention heads.D: Output dimension.

### 3.3 Comparison with baseline method

This paper compares the proposed Node2Vec-DGI-EL algorithm with several benchmark algorithms, including LAGCN, LBMFF, DRWBNCF [41], REDDA [42], KGCNH [43], and HTINet2. The experimental results are presented in Table 5. The results show that the Node2Vec-DGI-EL algorithm outperforms the other benchmark algorithms across multiple evaluation metrics. Specifically, the AUC reaches 0.9987 (Figure 3-A), and the AUPR reaches 0.9545 (Figure 3-B). Compared to the second-best model, DRWBNCF (AUC=0.991, AUPR=0.1136), the AUC improves by 0.78%, and the AUPR improves by 15.84%. LAGCN achieves an AUC of only 0.7998 and an AUPR of 0.1136, highlighting its limitations for this task. Models such as DRWBNCF and KGCNH also achieve good performance in AUC and AUPR but still show a certain gap compared to Node2Vec-DGI-EL. In summary, the Node2Vec-DGI-EL algorithm combines the graph embedding capability of Node2Vec with the representation learning advantage of the DGI framework, enabling it to effectively capture latent information in the data and exhibit superior predictive performance.

Table 5 Performance compared with 6 baseline methods.

| Model | AUC | AUPR | ACC | Precision | Recall | F1 |
|---|---|---|---|---|---|---|
| LAGCN | 0.7998 | 0.1136 | 0.7104 | 0.5265 | 0.7275 | 0.4709 |
| REDDA | 0.9324 | 0.4032 | 0.8948 | 0.5637 | 0.8763 | 0.6019 |
| LBMFF | 0.9631 | 0.4261 | 0.9241 | 0.5992 | 0.9114 | 0.6439 |
| HTINet2 | 0.9835 | 0.464 | 0.9186 | 0.5969 | 0.9253 | 0.6399 |
| KGCNH | 0.979 | 0.7159 | 0.9427 | 0.6315 | 0.9384 | 0.6524 |
| DRWBNCF | 0.991 | 0.7961 | 0.9508 | 0.6449 | 0.9577 | 0.7105 |
| **Node2Vec-DGI-EL** | **0.9987** | **0.9545** | **0.9827** | **0.7716** | **0.9835** | **0.8458** |

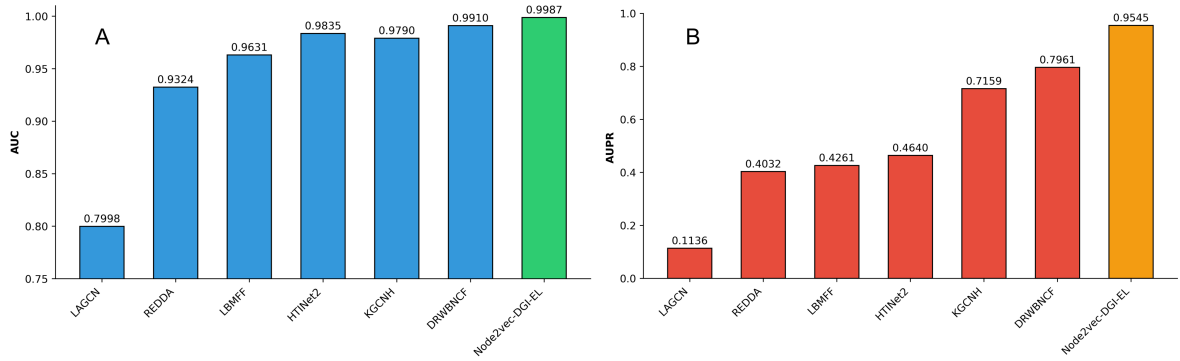

Figure 3 AUC and AUPR of 7 methods.

### 3.4 Ablation Studies

This section designs three ablation models to verify the effectiveness of each module in the Node2Vec-DGI-EL model: (1)Node2Vec-EL, which removes the DGI algorithm; (2)DGI-EL, which removes the Node2Vec algorithm and uses the node's one-hot encoding as the initial feature; (3)Node2Vec-DGI-EL, the complete model architecture. The experimental results are shown in Table 6. The AUPR of the complete model architecture is 0.9545, which is significantly higher than that of the other variants. The AUPR of Node2Vec-EL is 0.7833, with a performance decrease of 17.12%, indicating that the DGI module effectively improves the differentiation of node representations through unsupervised contrastive learning. The AUPR of DGI-EL is 0.8484, with a performance decrease of 10.16%, indicating that the node feature information extracted by Node2Vec enhances the representational capacity of the DGI algorithm. In conclusion, the synergy between the Node2Vec and DGI modules significantly improves model performance.

Table 6 Comparison of ablation test results.

| Model | AUC | AUPR | ACC | Precision | Recall | F1 |
| --- | --- | --- | --- | --- | --- | --- |
| Node2Vec-EL | 0.9867 | 0.7833 | 0.9357 | 0.6167 | 0.9406 | 0.671 |
| DGI-EL | 0.9943 | 0.8484 | 0.9581 | 0.6624 | 0.9659 | 0.733 |
| **Node2Vec-DGI-EL** | **0.9987** | **0.9545** | **0.9827** | **0.7716** | **0.9835** | **0.8458** |

### 3.5 Robustness Analysis

To evaluate the robustness of the Node2Vec-DGI-EL model, we simulated varying degrees of data loss by randomly hiding edges between ingredients and diseases at different proportions (ranging from 10% to 90%) and analyzed the model's performance under these conditions. The experimental results are shown in Figure 4. As the proportion of hidden edges gradually increases, the information that the model can obtain between ingredients and diseases decreases, leading to a gradual decline in the AUPR value. However, the model's AUPR value remains above 0.9, demonstrating that the Node2Vec-DGI-EL model can still effectively maintain high prediction accuracy and demonstrate strong robustness when facing varying levels of data loss.

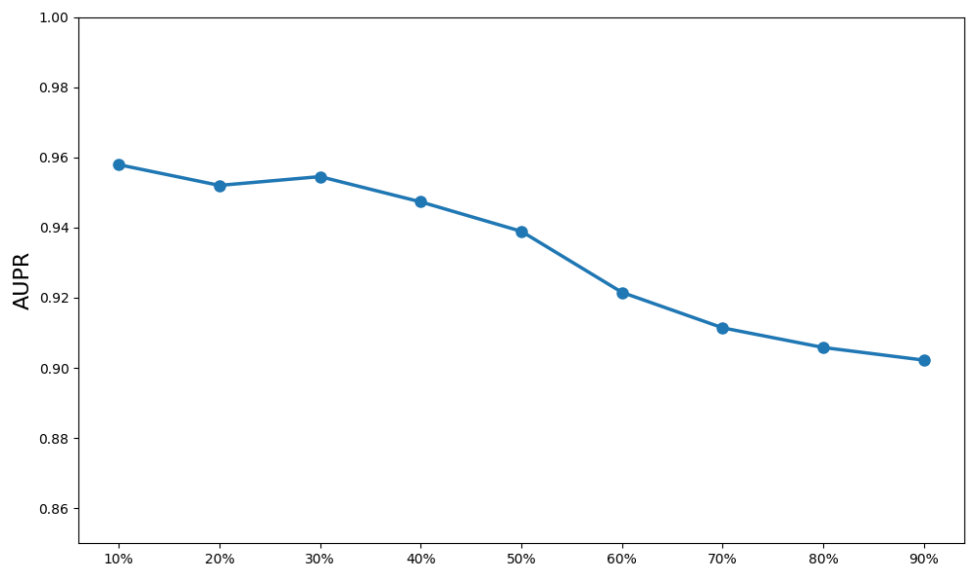

Figure 4 Model AUPR at different levels of data loss.

### 3.6 Case Analysis

This study analyzes two perspectives: predicting diseases highly correlated with ingredients and predicting Ingredients highly correlated with diseases, in order to comprehensively validate the reliability of the model's predictions.

### 3.6.1 Ingredient-disease association prediction

Triptonide (PubChem CID: 65411) was selected as an example. Based on the model's predicted values, diseases with no known association in the network were ranked in descending order (Table 7). Literature reviews have shown that triptonide can improve streptozotocin-induced diabetic retinopathy in rats [44], and that low-dose triptonide can effectively protect retinal cells from oxidative damage and inflammation [45], suggesting a potential relationship between triptonide and hypertensive retinopathy. However, no studies have indicated that triptonide was associated with the other four diseases, although it shared a high structural similarity with triptolide [46], which is linked to these diseases.

This study further explores the relationship between triptonide and hypertensive retinopathy. First, the action targets of triptonide and the disease targets for hypertensive retinopathy are identified, with 33 common targets being obtained through their intersection (Figure 5-A).We constructed a PPI network using STRING and performed visualization analysis with Cytoscape. The core targets were prioritized by the Maximal Clique Centrality (MCC) algorithm, revealing the top 10 hub targets (Figure 6-B). Subsequently, Triptonide was docked with the top 5 core target molecules using AutoDock to identify the optimal binding conformation and calculate molecular binding energy (Table 8).The results indicate that the molecular binding energy between triptonide and PGR is -9.62 kcal/mol, suggesting a stable interaction between the two. The visualization results are shown in Figure 5-C.

Table 7 Triptonide and the predicted top 5 diseases.

| Ingredient | Disease | Score | References |
|---|---|---|---|
| Triptonide | Hypertensive Retinopathy | 0.995 | [47],[48] |
| | Hypertensive Disease | 0.994 | [49] |
| | Insulin Resistance | 0.992 | [50] |
| | Accelerated Skeletal Maturation | 0.986 | [51,52,53] |
| | Acanthosis Nigricans | 0.986 | [54,55] |

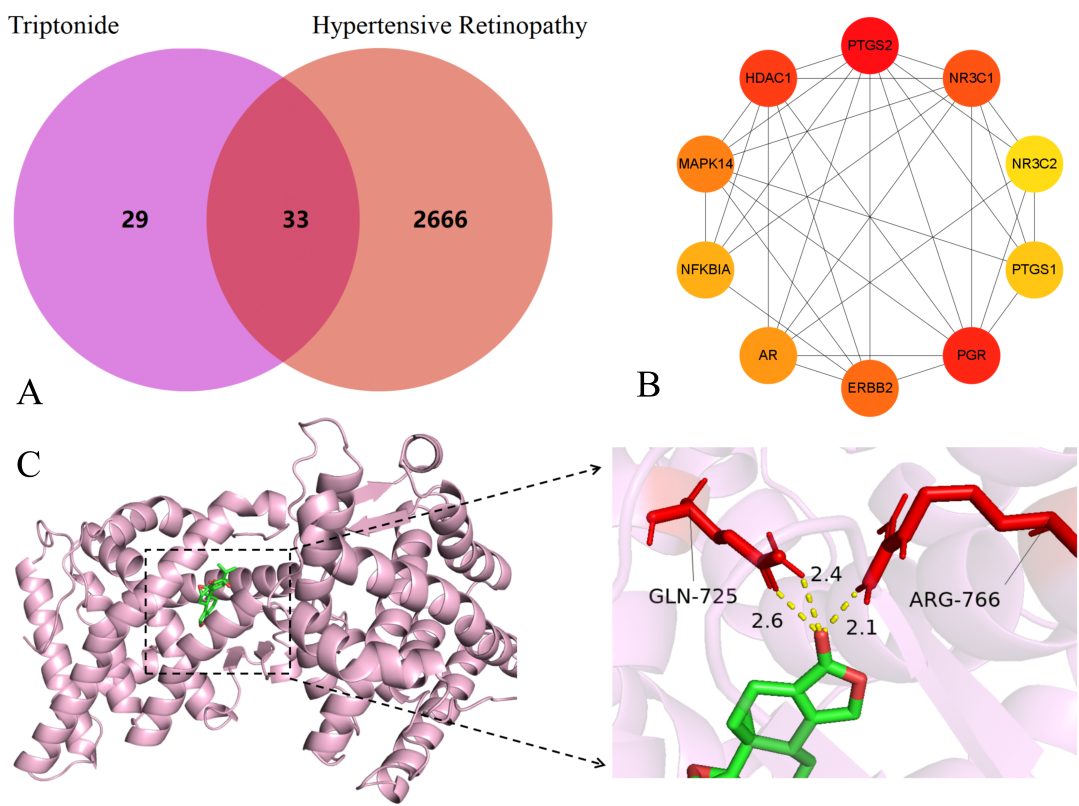

Figure 5　Analysis results of Triptonide and hypertensive retinopathy

Table 8　Molecular docking results.

| Target protein | PTGS2 | PGR | HDAC1 | NR3C1 | ERBB2 |
|---|---|---|---|---|---|
| Binding energy（kcal/mol） | -7 | -9.62 | -5.45 | -6.06 | -7.57 |

### 3.6.2 Disease-ingredient association prediction

This section presents colorectal cancer (CRC) as a case study. The model predicted a high association score (0.9966) between CRC and methyl ursolate (PubChem CID: 636516), suggesting significant pharmacological potential. We identify 52 shared targets through target intersection analysis (Figure 6-A). We constructed a PPI network using STRING and performed visualization analysis with Cytoscape. The core targets were prioritized by the MCC algorithm, revealing the top 10 hub targets (Figure 6-B).Molecular docking with AutoDock shows NFE2L2 has the strongest binding affinity (-9.71 kcal/mol) with methyl ursolate (Table 9, Figure 6-C), indicates stable ligand-receptor interaction.

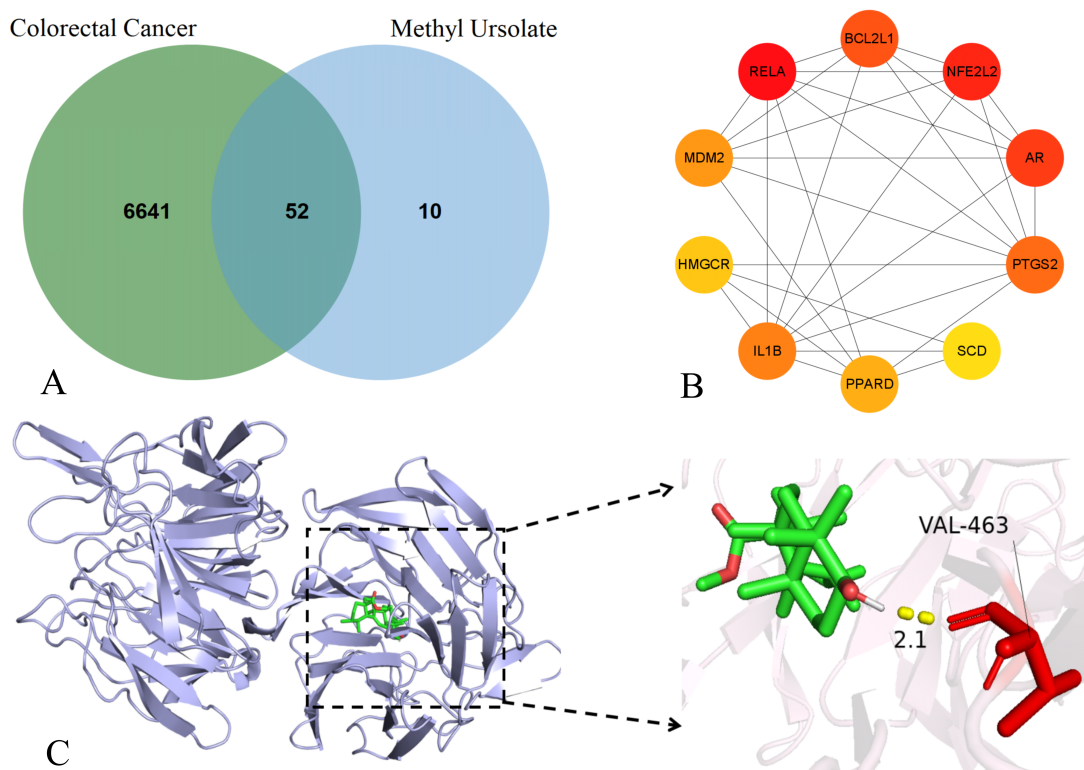

Figure 6 Analysis results of colorectal cancer and methyl ursolate.

Table 9 Molecular docking results.

| Target protein | RELA | NFE2L2 | AR | BCL2L1 | PTGS2 |
|---|---|---|---|---|---|
| Binding energy (kcal/mol) | -6.88 | -9.71 | -7.85 | -6.85 | -6.91 |

## 4. Discussion

TCM ingredients represent a vital source for drug discovery, serving as both the foundation for traditional drug development and a critical resource for modern pharmaceutical innovation. In this study, we first evaluated the predictive performance of five classical graph representation learning algorithms alongside the DGI algorithm on the HITD network. Notably, the Node2Vec-DGI-EL model outperformed all comparator methods across multiple evaluation metrics. Furthermore, hyperparameter optimization yielded an optimal parameter combination, achieving exceptional predictive performance (AUC = 0.9987, AUPR = 0.9545). Importantly, ablation experiments demonstrated that the synergistic interaction between Node2vec and DGI modules is pivotal to model enhancement, suggesting these methods complement each other's limitations and collectively improve generalization. Finally, robustness analysis confirmed the model's reliability under varying degrees of data sparsity.

To further validate the model's efficacy, we conducted case studies on triptonide and methyl ursolate, investigating their potential therapeutic roles in hypertensive retinopathy and CRC, respectively. For triptonide, our predictions identified hypertensive retinopathy as its highest-scoring association, with molecular docking revealing strong binding (ΔG = −9.62 kcal/mol) to PGR, a core target in this condition. Given that triptonide exhibits broad pharmacological activities (e.g., anti-inflammatory, anticancer, and immunomodulatory effects [46]) and PGR mediates neuroprotection and ocular blood flow regulation [48], our findings suggest triptonide may mitigate hypertensive retinopathy by targeting PGR, offering novel therapeutic insights. Similarly, methyl ursolate showed the highest predicted association with CRC, binding robustly (ΔG = −9.72 kcal/mol) to NFE2L2 (NRF2), a key regulator of oxidative stress [56]. Although NFE2L2 activation protects normal cells, its hyperactivation in tumors promotes proliferation and drug resistance [57]. Thus, methyl ursolate's known anticancer properties [58,59] may arise from NFE2L2-mediated modulation of oxidative stress pathways, potentially inhibiting tumor progression.

However, several limitations warrant consideration. First, the HITD network relies primarily on ETCM and STRING, whose limited coverage may constrain the model's predictive accuracy. Future studies should integrate additional TCM-related data to enhance network comprehensiveness. Second, while DGI demonstrated competitive performance, alternative contrastive learning algorithms may further optimize representation learning. Lastly, although molecular docking provided preliminary validation, in vitro (e.g., cell-based assays) and in vivo (e.g., animal models) experiments are essential to fully assess the model's translational potential.

## 5 Conclusion

The Node2Vec-DGI-EL model proposed in this study integrated three computational approaches: the Node2Vec algorithm for graph representation learning, the DGI algorithm for depth graph embedding, and ensemble learning methods. By applying this framework to the HITD network, we systematically explored potential associations between ingredients and diseases, consequently enabling the identification of highly relevant ingredient-disease pairs. This integrative approach offers significant advantages for TCM research. Node2Vec-DGI-EL was improved based on the characteristics of TCM, particularly enhancing its ability to reveal hidden relationships and accelerating the modernization of TCM.

**Abbreviations**
TCM Traditional Chinese medicine
DGI Deep Graph Infomax
HITD Herb-Ingredient-Target-Disease
PPI protein-protein interaction
AI artificial intelligence
DFS depth-first search
BFS breadth-first search
GAT graph neural network
RF Random Forest
CRC colorectal cancer

MCC Maximal Clique Centrality

# Code availability
The code is availability at https://github.com/wayfarer569/Node2Vec-DGI-EL.